\documentclass[sigconf]{acmart}

\setcopyright{none}
\newcommand\footnotetextcopyrightpermission[1]{}
\acmConference{}{}{}
\AtBeginDocument{%
  }

\usepackage{graphicx}
\usepackage{subcaption}
\usepackage{svg}
\usepackage[ruled]{algorithm2e}
\begin{document}

\title{Deep Reinforcement Learning for Minimum Zero-Forcing Sets}


\author{Steve Halley}
\email{shalley@villanova.edu}
\author{Maur\'{i}cio Gruppi}
\email{mgouveag@villanova.edu}
\orcid{0000-0001-7548-5012}
\affiliation{%
  \institution{Department of Computing Sciences, Villanova University}
  \city{Villanova}
  \state{Pennsylvania}
  \country{USA}
}

\begin{abstract}
  This paper explores the problem of finding the minimum zero-forcing set on undirected graphs and proposes an adapted machine-learning framework to solve the problem. The minimum zero-forcing set problem is a graph coloring problem where the color of an initial set of nodes propagates throughout a network. The set of nodes is zero-forcing if it forces all uncolored nodes to change color under the constraint of the color-change rule. There are several applications to this problem across different domains such as network science, network control, and designing logical circuits. Finding the minimum zero-forcing set is shown to be NP-hard. We propose a reinforcement learning framework, SD-ZFS, that adapts the S2V-DQN architecture to the ZFS problem. We train several models on this adapted framework and analyze the performance across graph datasets that have varying structures. We evaluate how the models trained on the framework generalize, scale, and transfer to different network types. The results demonstrate the effectiveness of the framework when compared against the optimal solution and greedy heuristic. We provide further insight into how the ZFS problem can be solved through machine-learning and the influence of network structure on the problem.
\end{abstract}




\maketitle
\fancyhead{}

\section{Introduction}
Zero-forcing is a propagation-based process over a graph where an initial set of nodes is colored and this set forces neighboring nodes to be colored in subsequent steps in accordance with the zero-forcing rule.  The zero-forcing rule states that a node changes color if and only if it is the only uncolored neighbor of a colored node. A zero-forcing set of nodes is a set of nodes that forces all uncolored nodes after an arbitrary number of iterations of the zero-forcing rule. The graph parameter $Z(G)$ represents the minimum size of a zero-forcing set of vertices. This parameter was initially proposed as a means to bound the minimum rank for numerous families of graphs \cite{aim2008zfs}. Further research has revealed a variety of applications such as the controllability of both quantum and classical systems\cite{burgarth2013zero}, monitoring electric grids \cite{haynes2002domination}, and designing logic circuits \cite{burgarth2014logic}. 

The problem of finding the minimum zero-forcing set has been shown to be NP-hard inspiring research into practical ways to find approximate solutions \cite{aazami2008hardness}. Several approaches have been taken to find an optimal solution faster than brute force such as the Wavefront algorithm \cite{butler2014minimumrank}, combinatorial algorithms and integer programming formulations \cite{brimkov2019computational}, and combining Boolean satisfiability modeling with a constraint generation framework \cite{brimkov2021improved}. Additionally, heuristics have been proposed to solve the problem in an iterative manner in polynomial time. \citet{brimkov2019computational} propose three variations of a greedy approach where nodes are iteratively added to the zero-forcing set based on the criteria of single vertex largest closure, neighborhood largest closure, or neighborhood scaled closure. These heuristics can be used to generate  solutions in polynomial time.

The time complexity of heuristic algorithms for the zero-forcing set problem varies depending on the density of the graph. For example, a simple greedy heuristic that iteratively selects the vertex producing the largest closure requires evaluating the closure operation for each candidate vertex at every iteration. Since a single closure computation can be performed in $O(n+m)$ time, the overall worst-case complexity of the greedy algorithm is $O(n^2*(n+m))$. This time complexity proves to be intractable on large graphs with high density leading to poor scalability. The performance of the heuristics can be close to optimal depending on the structure of the network but there is no known method of calculating the approximation ratio of the greedy solution size relative to the optimal solution size. 

The goal of the work presented in this paper is to apply a machine-learning approach to the problem of finding the minimum zero-forcing set through an adaptation of the S2V-DQN framework. 
Through this framework, we will analyze the performance of multiple models across several datasets of differing network structures. 
Unlike \citet{ahmad2024gml}, which formulate the problem as supervised node selection using a graph convolutional network trained to imitate greedy behavior, our approach formulates minimum zero-forcing set construction as a sequential decision-making problem and applies reinforcement learning through an adaptation of the S2V-DQN framework. This allows the model to optimize long-term rewards associated with solution quality rather than directly imitating a predefined heuristic.

This work is guided by the following research questions:

\begin{enumerate}
    \item Can the S2V-DQN framework generate a novel heuristic solution to the problem of finding the minimal zero-forcing set?
    \item How does the network structure affect the outcome of the ZFS problem?
    \item Will training the framework on specific network structures transfer to unseen data in a way that outperforms the greedy algorithm by taking advantage of certain topological properties of the network structure?
\end{enumerate}

Our contributions are:
\begin{enumerate}
    \item We have adapted the framework into our version denoted SD-ZFS (Structure2Vec-DQN-ZFS).
    \item We have produced several models of the framework and evaluated their performance against both greedy and approximate solutions. 
    \item We have found that the models have the capability of outperforming the greedy algorithm in certain cases and closely approximating the optimal solution. 
    \item We have analyzed the performance of the framework against a set of network properties for each set of test data, such as network density, clustering coefficient and network structure.
    \item We have demonstrated that the framework has the potential to generalize and scale to networks that vary in size and structure.
\end{enumerate}

\section{Background}

\subsection{Zero-Forcing Process and Relevant Algorithms}
An undirected graph $G = (V,E)$  represents a set of vertices and connecting edges. All pairs $\{u,v\} \in E$, connect nodes $u,v \in V$. A node u is a neighbor of node v if there is an edge $\{u,v\}$ in $E$. The neighborhood of a node $v$, denoted $N(v)$, is the set of all neighbors of $v$. The degree of node $v$, $d(v)$ is the cardinality of the neighborhood of $v$, $|N(v)|$. 

Given a graph, $G = (V,E)$, there is an initial set $S \subset V$ of vertices that are colored blue and the remaining vertices $T = V\textbackslash S$ are colored white. The zero-forcing rule dictates that if a node $v \in S$ has a single uncolored neighbor u, then $v \in S$ forces $u \in T$. The zero-forcing rule is implemented until it halts and no further nodes can be changed from white to blue. The result is a set of nodes, $B$, containing the initial set of nodes $S$ and each of the nodes that were forced. A minimum zero-forcing set is found when $|B|=|V|$.

\begin{figure}[h!]
    \caption{ZFS Propagation}
    \centering
    \includegraphics[width=\linewidth]{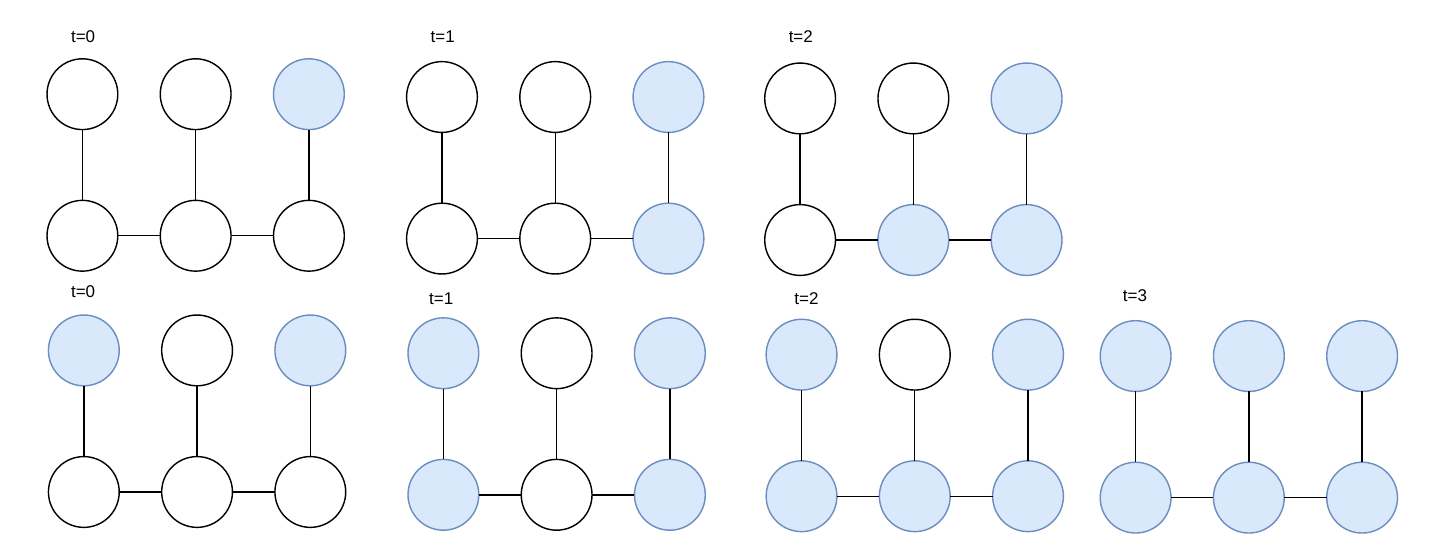}
    \label{fig:zfsprop}
\end{figure}

Figure 1 shows the zero-forcing process as it propagates from an initial set of blue nodes. Row 1 demonstrates the process as the coloring spreads from the top right node through two propagation steps before stalling. Row 2 shows that adding the node from the top left to the zero-forcing set leads to a forced network after three steps. This indicates that the top left and top right nodes form a zero-forcing set. Through brute-force, it is evident that any combination of two of the nodes in the top portion of the network will lead to a zero-forcing set but there is no single node that can force the entire graph blue. As a result, $Z(G)=2$ in Figure 1. 

Algorithms 1 and 2 demonstrate propagation and the greedy algorithm for solving the minimum zero-forcing set. The greedy algorithm operates by calculating the maximum size of the closure of the graph if a node is added to the zero-forcing set $Z$. At each iteration, the node with the highest gain is added to the set until a set that forms a closure of the full network is found.

\begin{algorithm}[h]
\caption{Zero-Forcing Color-Change Rule}
\label{alg:zfrule}
\KwIn{Graph $G=(V,E)$, initial blue set $S \subseteq V$}
\KwOut{Final closure $\operatorname{cl}(S)$}

Mark all vertices in $S$ as blue\;
Mark all vertices in $V \setminus S$ as white\;

\Repeat{$changed = \textbf{false}$}{
    $changed \gets \textbf{false}$\;
    
    \ForEach{blue vertex $u \in V$}{
        $W \gets \{v \in N(u) : v \text{ is white}\}$\;
        
        \If{$|W| = 1$}{
            Let $v$ be the unique vertex in $W$\;
            Color $v$ blue\;
            $changed \gets \textbf{true}$\;
        }
    }
}

\Return{set of blue vertices}
\end{algorithm}

\begin{algorithm}[h]
\caption{Greedy Zero-Forcing Set Heuristic}
\label{alg:greedyzfs}
\KwIn{Graph $G=(V,E)$}
\KwOut{Approximate zero-forcing set $Z$}

$Z \gets \emptyset$\;
$B \gets \emptyset$\;

\While{$|B| < |V|$}{
    $best\_v \gets \text{null}$\;
    $best\_gain \gets -1$\;
    
    \ForEach{$v \in V \setminus Z$}{
        $B' \gets \operatorname{cl}(Z \cup \{v\})$\;
        $gain \gets |B'|$\;
        
        \If{$gain > best\_gain$}{
            $best\_gain \gets gain$\;
            $best\_v \gets v$\;
        }
    }
    
    $Z \gets Z \cup \{best\_v\}$\;
    $B \gets \operatorname{cl}(Z)$\;
}

\Return{$Z$}
\end{algorithm}

\subsection{S2V-DQN}

Existing greedy heuristics such as Algorithm 2 for approximating minimum zero-forcing sets are typically based on local, immediate gains. These approaches often fail to incorporate richer structural information about the graph and the role of individual vertices within that structure. We propose that learning representations of graph topology and node-specific properties can provide additional information for guiding iterative node selection toward smaller zero-forcing sets. 

A common approach for learning on graph-structured data is to use Graph Neural Networks (GNNs), which learn representations that capture both local node features and global graph structure. The representations can be combined with reinforcement learning to generate a framework in which the embedding process is integrated with a specific problem that the network is trained on. The hybrid framework proposed by \citet{dai2017dqn} in the context of combinatorial optimization problems is S2V-DQN which combines graph representation learning(Structure2Vec) with deep reinforcement learning(Deep Q-Learning Network). S2V-DQN is a two component end-to-end framework designed to automate the process of finding heuristic algorithms for NP-hard combinatorial optimization problems that exploits the structure of recurring problems\cite{dai2017dqn}. 
Prior work has successfully learned heuristic policies to learn greedy heuristics on similar graph theoretic problems such as Minimum Vertex Cover, Max-Cut, and Traveling Salesperson. 
\citet{dai2017dqn} demonstrated a close approximation ratio to optimal after training S2V-DQN on instances of these problems. For example, the S2V-DQN framework demonstrated the ability to find the solution to MVC on a 1200 node graph in 11 seconds on a single GPU with an approximation ratio of 1.0062. These results demonstrate a promising scalability and trade-off in solution quality to optimal. 

The framework operates through a graph embedding network that processes graph-structured data by iteratively encoding node representations through message passing. This iterative procedure allows each node embedding to capture information about both the node itself and the structural context of its local neighborhood. One approach for inferring latent representations of a node’s role within a graph is mean-field inference  where node embeddings are iteratively updated based on neighboring embeddings\cite{dai2017dqn}. A node $v$ is initially parameterized by an embedding vector $\mu_v^{(t=0)}$ which is then recursively updated after $t$ message-passing iterations as shown below. At iteration $t+1$, $W_1$ and $W_2$ represent learnable weight matrices associated with the node features $x_v$ and the neighboring node embeddings $\mu_u^{(t)}$.

\begin{equation}
    \mu_v^{(t+1)} = \sigma ( W_1 x_v + W_2 \sum_{u \in \mathcal{N}(v)} \mu_u^{(t)} )
\end{equation}

Edge information can be captured with an additional term:
\begin{equation}
\mu_v^{(t+1)}
=
\sigma \left(
W_1 x_v
+
W_2 \sum_{u \in \mathcal{N}(v)} \mu_u^{(t)}
+
W_3 \sum_{u \in \mathcal{N}(v)}
\sigma \left( W_4 w(v,u) \right)
\right)
\end{equation}

$\sigma$ denotes the rectified linear unit activation function. After $t$ hops, local information about the $t$-hop neighborhood is encoded. The embeddings can then be aggregated with sum pooling (Equation 3) or mean pooling (Equation 4) to construct a global representation of the graph.

\begin{equation}
    H_G = \sum_{v \in V} \mu_v
\end{equation}

\begin{equation}
    H_G = \frac{1}{|V|} \sum_{v \in V} \mu_v
\end{equation}

Other message-passing approaches have been extensively explored in the graph neural network literature. For further reading,\citet{yedidia2002understanding} discuss belief propagation and its generalizations where inference is performed through iterative local message passing between neighboring nodes.

The second half of the framework(DQN) is a Deep Q-Learning Network. Originally proposed by researchers at DeepMind\citet{mnih2015human}, DQN builds on the concept of q-learning which is a reinforcement learning approach to problem-solving where an action is selected based on the q-score. A q-function  transforms information about the current state into a quantitative evaluation of each possible action. The q-function is trained over a series of episodes of a specific problem. DQN builds on this idea by utilizing a neural network to approximate the optimal action-value function:

\begin{equation}
    Q^*(s,a) = \max_\pi \mathbb{E} \left[ \sum_{k=0}^\infty \gamma^k r_{t+kj} \vert s_t = s, a_t=a\right]
\end{equation}

The function takes the current state, and evaluates the action based on the discounted expected reward according to behavior policy $\pi$. In the context of graph theoretic combinatorial optimization problems, this equation can be transformed into an approximate q-score function ${Q}(h(S), v; \Theta)$\cite{dai2017dqn}. This function evaluates a q-score based on the current state of the graph($h(S)$) and the vertex added($v$). In place of the vertex and graph, we substitute the node embedding and graph embedding derived from the S2V part of the framework:

\begin{equation}
\\Q(h(S), v; \Theta)
=
W_5
\,
\mathrm{ReLU}
\left(
\left[
W_6 \sum_{u \in V} \mu_u^{(t)},
\;
W_7 \mu_v^{(t)}
\right]
\right)
\end{equation}

This process of capturing node and graph representations through S2V, and then using the representations as inputs for the DQN to approximate q-scores creates a framework that can be applied to a variety of combinatorial optimization problems. The approximate q-function that is learned through training the network on features specific to a problem is designed to work in a similar manner to a greedy heuristic by greedily adding nodes to the solution that have the highest q-score. 

Building on the S2V-DQN framework described above, we train neural networks tailored to the ZFS problem to iteratively select nodes that construct a minimum zero-forcing set. We hypothesize that the framework has the potential to learn new greedy heuristics through the q-function training process that can lead to novel solution strategies by picking up on structural network properties.

\subsection{Network Families}
Graphs can be broadly categorized into many different families that share similar characteristics.  It can often be useful for graph theoretic problems to understand the structural properties of the network. This often leads to more efficient solutions on certain types of networks for a specific problem. For example, Minimum Vertex Cover can be solved in polynomial time for bipartite graphs through the Hopcroft–Karp algorithm\cite{hopcroft1973n}. 

The ZFS problem closely relates to the  structure of the graph. The size of the minimum zero-forcing set can be as large as $|V-1|$ on complete graphs\cite{aim2008zfs} or as small as 1 for a straight line by picking an endpoint of the line of nodes. One important structure is a tree $T = (V,E)$ where T is connected and acyclic. In a tree, there is only a single path between any pair of vertices. It has been shown that the minimum zero-forcing set of a tree is directly related to the minimum path cover number such that $M^F(T) = Z(T)$\cite{aim2008zfs}. The zero-forcing set can be found in this case by selecting one endpoint in each path of the minimum path cover. In this specific case, the time to solve for the ZFS is equivalent to the the minimum path cover which can be found in polynomial time $O(N)$. We hypothesize that S2V-DQN may be able to exploit structural properties similar to those mentioned above to generate new solution strategies. 

Two key categories of networks that share structural similarities across scale are Random Networks and Scale-Free networks. 

Random networks were designed to model properties of real-world networks by linking isolated nodes if a randomly generated number exceeds a p-value set to a number between 0 and 1. When density is kept consistent over the generation of numerous networks, a similar structure will appear across the samples that includes a binomial degree distribution, independent local clustering coefficients, and a low standard deviation from average degree\cite{barabasi2016network}. There is also a lack of hubs across random networks due to the fact that no node has any greater probability than another of connecting to a large number of nodes. 

Scale-free networks have a degree distribution that differs from  random networks. The degree distribution follows a power-law resulting in a large tail on the distribution of high-degree nodes\cite{barabasi2016network}. This results in a large number of hubs surrounded by low-degree nodes. 

In later sections we will train our framework on networks from both of these families of networks and see what kind of impact the type of training data has on inference across datasets of different varieties of networks such as selecting a model trained on scale-free networks and testing it on real-world networks. 

\subsection{Related Approaches}

A similar approach has been made to solve for the minimum zero-forcing set through the application of a Graph Convolution Network (GCN) by \citet{ahmad2024gml}. Their approach to the GCN for ZFS is designed to imitate the greedy algorithm while achieving significant speed improvements to surpass the computational efficiency of the greedy algorithm. This research showed it is feasible to solve the ZFS problem with a GNN approach and they also showed improved solution quality over certain data and a computation time that surpassed the greedy heuristic by several orders of magnitude. They also contributed a rich network dataset containing the greedy solution to the ZFS problem. We will utilize several sets of these networks in our evaluation of our proposed framework and also compare our results on a limited subset of data. 

\section{S2V-DQN-ZFS Architecture}
In this section we will discuss the adaptation of S2V-DQN to the problem of finding the minimum ZFS of a network. In the following sections we will cover the overall structure of the adaptation and the approach taken to train several models on the architecture. The adapted framework will be denoted as SD-ZFS going forward which synthesizes Structure2Vec graph representation learning(S2V) and Deep Q-Learning networks(DQN) with the constraints of the zero-forcing rule and the goal of finding the minimum zero-forcing set(ZFS). 
\subsection{Structure}
SD-ZFS is an end-to-end framework designed to solve the ZFS problem by embedding the network and nodes of a graph and transforming the embedding into a scalar value that represents q-scores for each individual node relative to their quantitative value towards finding the minimum zero-forcing set. 

First we formulate the ZFS problem as a reinforcement learning environment:

\begin{itemize}
    \item State: Information about the graph and current partial solution(nodes selected for the zero-forcing set).
    \item Action: An action in the context of the problem is any node selected to be added to the zero-forcing set .
    \item Transition: Information about action taken and results including reward and following state.
    \item Reward: A reward of -1 is added for each additional node to the zero-forcing set.
    \item Policy: Deterministic greedy policy that selects an action based on the estimated q-score(See Equation 7).
\end{itemize}

Initially all nodes are set to white and the partial solution will be an empty set. This environment is where the SD-ZFS is trained and evaluated. The implementation pseudo-code is detailed in Algorithm 3.

After initialization, each node feature vector(representative of individual nodes in the graph) forms the input to the neural network. 

\begin{algorithm}[h!]
\caption{ZFSEnv}

\textbf{Initialize Environment()}\;
Store graph information and compute node features. $S$ = set of all selected nodes, $B$ = set of all selected nodes and all nodes forced blue\;

\textbf{Reset()}\;
Clear selected nodes and initialize state/features\;

\textbf{Legal Actions($S$)}\;
Return all nodes not yet selected\;

\textbf{Closure($S$)}\;
Apply the zero-forcing rule until no more nodes can be forced (Algorithm 1)\;

\textbf{Zero Force Propagation($S$)}\;
Update set of blue nodes using Closure\;

\textbf{Done($S$)}\;
Check if all nodes are blue\;

\textbf{Prune($S$)}\;
Remove redundant selected nodes while preserving ZFS\;

\textbf{Step(action)}\;
Select a node, propagate forcing, compute reward, and return next state\;

\end{algorithm}

We then construct a two-component neural network representing the primary components of S2V-DQN. Mean-field inference, as described in Equation 1, is utilized to iteratively embed each node into an $l$-dimensional latent representation, where $l$ denotes the latent dimension and the embeddings are initialized from the node feature vectors. The resulting node embeddings are aggregated through sum pooling, as described in Equation 3, to generate a graph-level embedding. The concatenation of the node embedding and graph embedding forms the output of the first component, which is designed to represent both the local structure of a node and the global structure of the graph within the context of the current problem state. 

The second component of the neural network consists of a multi-layer perceptron(MLP) that takes the concatenated node and graph embeddings as an input. The embeddings are passed through an $h$-dimensional hidden layer with ReLU activation before being mapped to a final scalar q-score output. 

The q-score represents the estimated value of adding that specific node to the current partial solution. A deterministic greedy policy
\begin{equation}
\pi(v \mid S) := \arg\max_{v' \notin S} Q(h(S), v')    
\end{equation}
is used to select the next action\cite{dai2017dqn}. This node gets added to the environment through the step function and  the graph is checked for closure with the new partial solution. The environment will then calculate legal actions(based on all unselected nodes), re-calculate node feature, provide a reward, and return the new state. This data gets bundled in a transition which is used to train the network. This process continues until the framework finds a solution to the problem. 

In the case where a solution is found, the selected nodes of the solution will be "pruned" where any redundancy is accounted for. This will remove any extraneous nodes from the set that are unnecessary to the forcing process and ultimately are forced blue through propagation.

\begin{algorithm}[h!]
\caption{SD-ZFS Training}
\label{alg:s2vdqn-training}
\KwIn{Graph distribution $\mathcal{D}$, replay memory capacity $N$, training episodes $K$, batch size $B$, discount $\gamma$, target update frequency $C$, $n$-step horizon}
\KwOut{Trained parameters $\Theta$}

Initialize replay memory $\mathcal{M}$ to capacity $N$\\
Initialize policy network $Q(h(S),v;\Theta)$\\
Initialize target network $Q(h(S),v;\Theta^-)$ with $\Theta^- \leftarrow \Theta$\

Sample graph $G \sim \mathcal{D}$ and reset environment to obtain $(S_1,X_1)$\\
Initialize empty $n$-step buffer $\mathcal{B}_n$\\
Initialize state ($r$=0,$S_t$= $\emptyset$,$X_t$,done=False)\\
Let t=0\\
\For{$k = 1$ \KwTo $K$}{
    \While{not done}{\[
    v_k =
    \begin{cases}
    \text{random legal node } v \notin S_t, & \text{w.p. } \epsilon,\\
    \arg\max_{v \notin S_t} Q(h(S_t),v;\Theta), & \text{otherwise}.
    \end{cases}
    \]

    Execute $v_k$ and observe $(r_k,S_{t+1},X_{t+1},done)$\;
    Add transition to $\mathcal{B}_n$\

    \If{$|\mathcal{B}_n| \geq n$ }{
        add $\mathcal{B}_n$\ to $\mathcal{M}$\
    }

    Sample $B$ transitions $\mathcal{T}$from $\mathcal{M}$\;
    Update $\Theta$ by minimizing
    \[
    \mathcal{L}(\Theta)
    =
    \frac{1}{B}
    \sum_{\mathcal{T}}
    \mathrm{Huber}
\left(
Q(h(S_t), v_t; \Theta_t),
Y_t^{\mathrm{DDQN}}
\right)
    \]

    \If{$k \bmod C = 0$}{
        $\Theta^- \leftarrow \Theta$\;
    }

    Set $(S_t,X_t) \leftarrow (S_{t+1},X_{t+1})$\;

    \If{done}{
        Sample new graph $G \sim \mathcal{D}$ and reset environment\\
        Clear $\mathcal{B}_n$\\
        k++\\
        let t=0
    }}
    
}

\Return{$\Theta$}\;
\end{algorithm}

\subsection{Training}

Our approach to training largely mirrors the approach elaborated by Dai et al. with minor modifications. We initialize a policy  network $Q(h(S),v;\Theta)$ and a target network $Q(h(S),v;\Theta^-)$ with the same parameters. The policy network is utilized to select actions and trained continuously. The target network gets updated periodically after a stated number of training steps. The policy follows an epsilon-greedy approach to action selection where the policy has a probability 1-$\epsilon$ of selecting an action based on the q-score. Otherwise, the policy network will select a random action. This allows for an early series of random actions to be used for training. The training process is designed to decay epsilon over a number of training steps and the policy network will select an action based on q-score with increasing probability as the value of epsilon drops.

Each action and corresponding reward are stored as transitions in replay memory $\mathcal{M}$ along with relevant information including current/next node-features and state, reward, and if a solution was found. We utilize n-step transitions, where n controls the number of environment steps aggregated into a single transition.

After an initial warm-up period, the training loop samples minibatches of n-step transitions from replay memory $\mathcal{M}$ to calculate the loss and update the policy network parameters. We utilize the Double DQN (DDQN) approach for target computation, where the policy network selects the maximizing action and the target network evaluates the corresponding q-value. This separates action selection from action evaluation and helps reduce the overestimation bias commonly observed in standard DQN methods\cite{vanhasselt2016double}. Similar to standard Q-learning, the network parameters are updated by minimizing the Bellman error between the predicted q-score and the target q-score. Rather than utilizing a 1-step squared loss objective, we utilize DDQN with n-step returns and Smooth L1 (Huber) loss for improved stability. We follow the calculations for the target used by the DQN  proposed by Hasselt et al. where selection of action still use the parameters from the online policy but the parameters from the target policy are used to evaluate the value of the online policy\cite{vanhasselt2016double}:

\begin{equation}
Y_t^{\mathrm{DDQN}}
=
R_{t+1}
+
\gamma
Q
\left(
h(S_{t+1}),
\arg\max_{v}
Q(h(S_{t+1}), v; \Theta_t);
\Theta_t^{-}
\right)
\end{equation}

This target value is then used to calculate the loss between the policy generated q-value and target:

\begin{equation}   
\mathcal{L}(\Theta_t)
=
\mathrm{Huber}
\left(
Q(h(S_t), v_t; \Theta_t),
Y_t^{\mathrm{DDQN}}
\right)
\end{equation}

We utilize Huber Loss to better handle the loss from outliers. The policy network is then optimized through stochastic gradient descent. This process continues for a number of training steps and a new graph is selected for training after a solution is found. 

We utilized this training process detailed above and outlined in Algorithm 4 to train three models on three different sets of data. Each of the models is outlined below along with limited selected hyper-parameters in Table 1. The first model is trained on randomly generated Erdős–Rényi graphs with the NetworkX Python library with a p-value of 0.2 and 30 nodes. A new network is generated for each training episode. Similarly, the second model is trained on randomly generated Barabási–Albert networks with the NetworkX Python library with 20 nodes and an m-value of 2. The third model is trained on data randomly sampled from the Facebook dataset which is described in more detail in section 4.2. 
\begin{table}[h!]
\centering
\caption{Summary of trained SD-ZFS models and training configurations}
\scriptsize
\setlength{\tabcolsep}{4pt}
\begin{tabular}{llccccc}
\toprule
Model & Training Data & Network Type & Episodes & N-Steps & Latent & Hidden \\
\midrule
SD-ZFS ER & ER Networks & Random & 5,760 & 3 & 128 & 256 \\
SD-ZFS BA & Barabási--Albert & Scale-free & 6,765 & 3 & 128 & 256 \\
SD-ZFS FB & Facebook & Real World & 12,849 & 3 & 128 & 256 \\
\bottomrule
\end{tabular}
\label{tab:model_summary}
\end{table}

\section{Results}
\subsection{Evaluation}
Our evaluation focuses on three categories of networks: random, scale-free, and real-world. We will compare and contrast the performance of our models in each category of data to better understand how the structure of the training data will affect the model performance. We also seek to understand how well models trained on one type of network might generalize to others, and how these models scale to larger graphs. 

For a given dataset $S$, The key metric for evaluation will be the mean difference from the greedy algorithm and mean deviance from the greedy algorithm defined as: 

\begin{equation}
    Diff(S) = \frac{1}{|S|} \sum_{i=0}^{|S|}Z_{SD-ZFS}(G_i)- Z_{Greedy}(G_i)
\end{equation}
\begin{equation}
    Dev(S) = \frac{1}{|S|} \sum_{i=0}^{|S|} \frac{{Z_{SD-ZFS}(G_i)- Z_{Greedy}(G_i)}}{Z_{Greedy}(G_i)} *100
\end{equation}

Where applicable, we will also measure the mean approximation ratio to the optimal solution size as:

\begin{equation}
    Approx(S) = \frac{1}{|S|} \sum_{i=0}^{|S|}\frac{Z_{SD-ZFS}(G_i)}{Z_{Optimal}(G_i)}
\end{equation}

\subsection{Datasets}
The datasets utilized are partially taken from the contribution by \citet{ahmad2024gml} which contain a large sample of real-world and synthetic networks with the greedy solution size.

In addition, we have generated a few synthetic datasets to analyze the scalability of the framework. Further, to generate a sample  of real-world networks with limited size, we have taken a dataset of Facebook data containing user friends lists collected from surveys and used depth-first traversal from random nodes to create a dataset of small real-world networks.

\begin{table}[h]
\caption{Summary of graph datasets}
\centering
\resizebox{\columnwidth}{!}{
\begin{tabular}{lcccccc}
\hline
Graph Dataset & \# of graphs & Range ($|V|$) & Avg ($|V|$) & Avg Degree & Avg Greedy($G$) \\
\hline
Small ER*        & 979  & 30--70    & 65    & 12.86 & 40.56 \\
Large ER*        & 1500 & 500--1000 & 744.8 & 19.64 & 448.11 \\
Large Barabási–Albert   & 1000 & 31--998   & 500.36 & 29.47  & 284.05 \\
Small Barabási–Albert   & 1000 & 15-30   & 22.5 & 2.78  & 9.01 \\
COLLAB*          & 5000 & 32--492   & 74.49 & 37.37 & 57.89 \\
IMDB-BINARY*     & 1000 & 12--136   & 19.7  & 8.89  & 16.13 \\
REDDIT-BINARY*   & 1834 & 6--1194   & 288.1 & 2.34  & 177.7 \\
Facebook   & 912 & 16--64  & 32.69 & 18.00  & N/A  \\
ER\_Scale        & 400  & 25--1000  & 379.38 & 75.72 & 346.43 \\
BA\_Scale        & 400  & 25--1000  & 379.38 & 7.76   & 102.20    \\
\hline
\end{tabular}
}
\vspace{2pt}
\raggedright
\footnotesize{ * denotes dataset contributed by Ahmad et al.}
\end{table}

\subsection{Random Networks}
We test our approach for random networks on three Erdős–Rényi network datasets. The baseline dataset is Small\_ER where the solution quality can be compared to both the optimal and greedy solution. 

Figure 2 shows the comparison between the performance of SD-ZFS ER on the dataset and the performance of the greedy heuristic. The model shows a close approximation to optimal on this dataset and a significant improvement in solution quality over greedy. The histogram in Figure 3 shows the observed gain in model performance over the greedy heuristic for the graphs within the dataset. The model matches or improves upon the greedy heuristic on every sample with a mean difference of 4.47 and a mean deviance of 11.45\%. This data shows that the model is outperforming the greedy algorithm by over 4 nodes on average for each of the samples. The approximation ratio of 1.048 indicates the model is closely approximating the optimal solution. In Figure 2, the blue bar shows the approximation ratio of the model to the optimal solution and the orange bar shows the approximation ratio of the greedy heuristic to the optimal solution. 
\begin{figure}[h!]
    \caption{Approximation Ratio of SD-ZFS ER against Greedy Heuristic}
    \centering
    \includegraphics[width=1\linewidth]{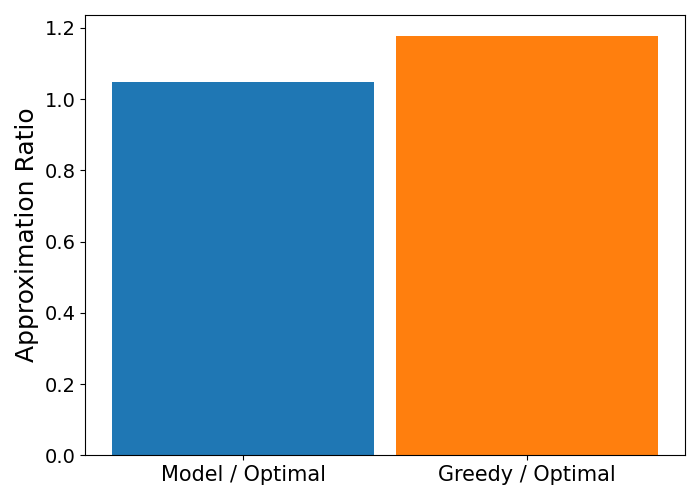}
    \label{fig:approxratio}
\end{figure}
\begin{figure}[h!]
\caption{Histogram showing SD-ZFS ER improvement on Small\_ER Dataset}
    \centering
    \includegraphics[width=1\linewidth]{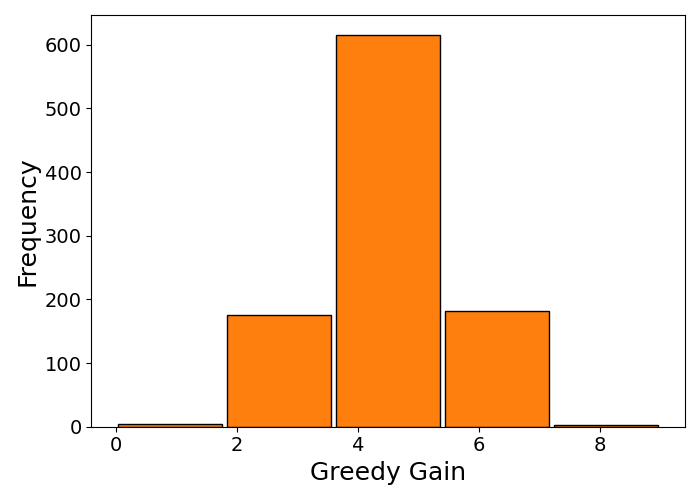}
    \label{fig:histogram}
\end{figure}

We tested the scalability of SD-ZFS ER on the ER\_Scale dataset to see if performance would be consistent as networks increased in size. The structure of the networks in this dataset is designed to be roughly uniform. The only variable that changes is the number of nodes so that the model performance can be closely interpreted with respect to scale.  The model shows continued performance gains over the greedy heuristics as network size scales from 25 nodes to 1000 nodes. As shown in figures 4 and 5, the model outperforms the greedy heuristic at each increment. 
\begin{figure}[h!]
\caption{ER Scale Results - SD-ZFS ER Comparison to Greedy}
    \centering
    \includegraphics[width=1.0\linewidth]{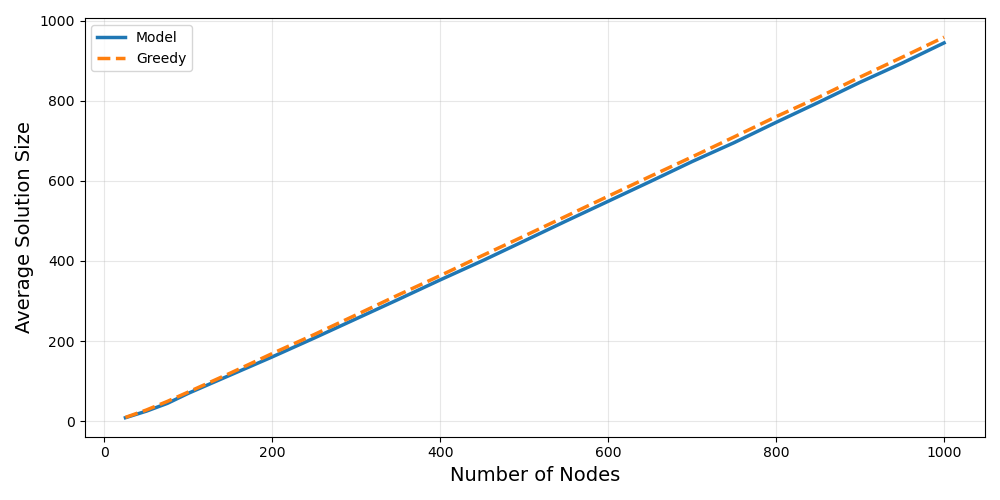}
    
    \label{fig:erscale}
\end{figure}
\begin{figure}
\caption{ER Scale Results - Enhanced View}
    \centering
    \includegraphics[width=1\linewidth]{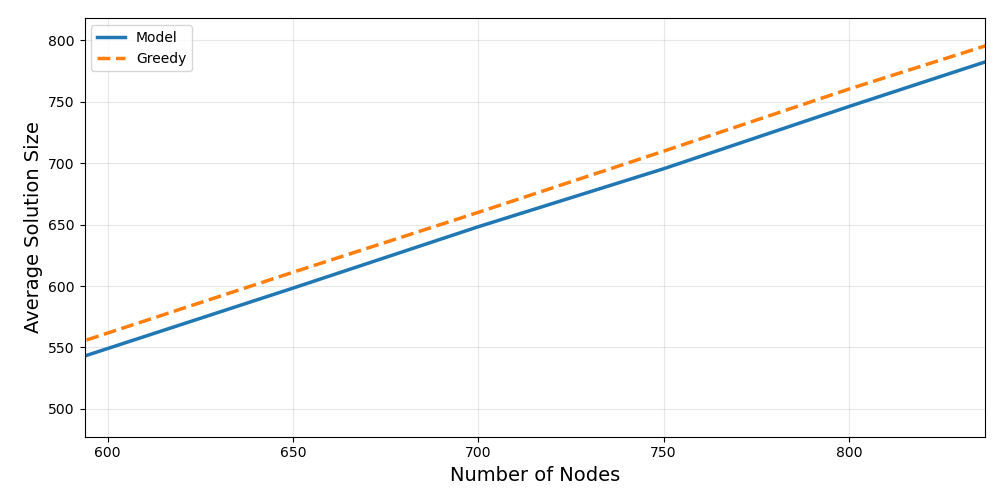}
    \label{fig:erscale_zoom}
\end{figure}

On the Large\_ER dataset, SD-ZFS ER performs slightly better than the greedy heuristic with a mean variance of 1.37 and a mean deviation of 0.71\% showing the capability to generalize and scale to networks of varying size and density. The win rate was 51.0\%, win/tie rate 54.0\%, and the loss rate 46.0\%. 

SD-ZFS BA performed the best out of our group of models with a mean difference of 1.24 and mean deviance of 1.18\%. See below for a comparison of the wins/losses against greedy for SD-ZFS BA.
\begin{table}[h!]
\caption{SD-ZFS BA results on Large ER dataset}
\centering
\begin{tabular}{lc}
\hline
\textbf{Outcome} & \textbf{Percentage} \\
\hline
Win     & 81.8\% \\
Win/Tie & 85.2\%  \\
Loss    & 14.8\%  \\
\hline
\end{tabular}
\label{tab:ba_large_er}
\end{table}

In summary, SD-ZFS ER outperformed the greedy algorithm on all random network datasets, and showed a promising approximation ratio on the Small\_ER dataset. Further data in Table 8 shows performance of models trained on other datasets, such as scale-free and real-world networks and how they perform on the random network datasets. 

\subsection{Scale-free Networks}

We tested our approach across three scale-free datasets including Small Barabási-Albert, Large Barabási-Albert, and  BA\_Scale. 

Figure 6 shows the comparison between the performance of SD-ZFS BA and the greedy heuristic on the Small Barabási-Albert dataset. Both the model, and the greedy heuristic, closely approximate the optimal solution. The model shows mild performance gains over greedy with a mean difference of 0.03 and mean deviation of 0.14\%. 

\begin{figure}
\caption{Approximation Ratio of S2V-DQN against Greedy
Heuristic }
    \centering
    \includegraphics[width=1.0\linewidth]{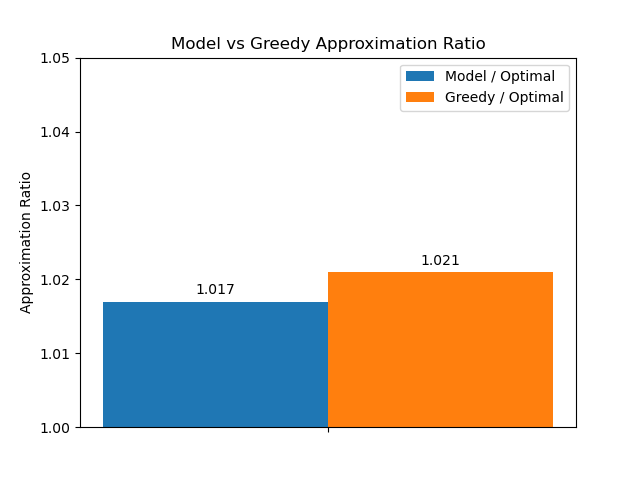}
    \label{fig:ba_small}
\end{figure}

SD-ZFS BA is tested on the BA\_Scale dataset to analyze performance over increasing network sizes. Similar to the ER\_Scale dataset, the BA\_Scale dataset was generated in a way to preserve structure across an increasing number of nodes. Figures 7 and 8 shows the results of the model against the greedy heuristic across the dataset. The result is a consistent out-performance of the greedy heuristic as networks grow increasingly larger. Similar to ER\_Scale, we note a strong scalability of the model when applied to networks that were similar in structure to that of the training data resulting in a strong performance relative to the greedy heuristic. 
\begin{figure}[h!]
\caption{BA Scale Results - SD-ZFS BA Comparison to Greedy}
    \centering
    \includegraphics[width=1\linewidth]{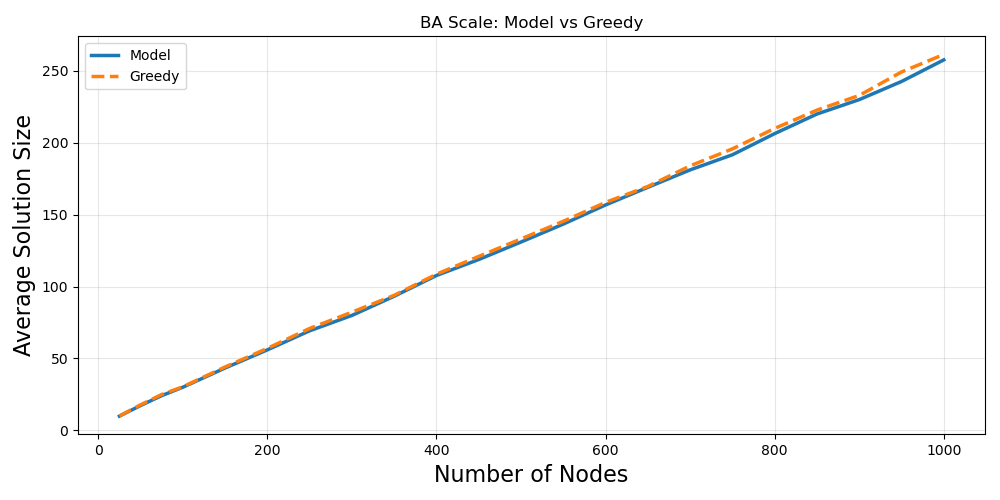}
    \label{fig:ba_scale}
\end{figure}
\begin{figure}
\caption{BA Scale Results - Enhanced View}
    \centering
    \includegraphics[width=1\linewidth]{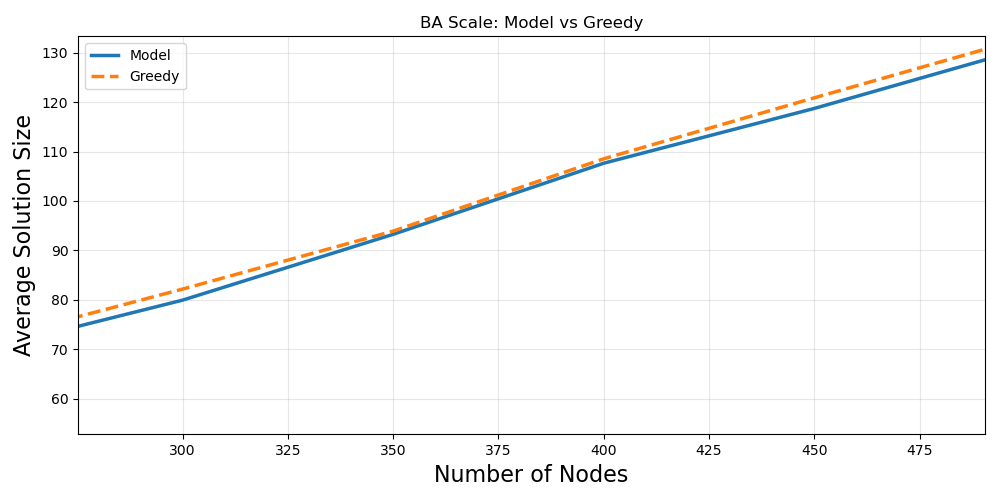}
    \label{fig:ba_zoom}
\end{figure}

The last dataset evaluated in the context of scale-free networks is Large Barabási–Albert. SD-ZFS BA shows a slight increase in performance over the greedy heuristic with a mean variance of 1.24 and a mean deviation of 1.128\%. 

\begin{table}[h!]
\caption{SD-ZFS BA results on Large BA dataset}
\centering
\begin{tabular}{lc}
\hline
\textbf{Outcome} & \textbf{Percentage} \\
\hline
Win     & 58.0\% \\
Win/Tie & 68.0\%  \\
Loss    & 32.0\%  \\
\hline
\end{tabular}
\end{table}

Overall, SD-ZFS BA showed it was capable of generalizing and scaling to larger networks of various densities. It was also able to surpass the performance of SD-ZFS ER on the Large ER dataset indicating that it could generalize well to random networks. 

\subsection{Real World Networks}
The third part of the evaluation focuses on real-world datasets including the COLLAB, IMDB-BINARY, and REDDIT-BINARY datasets. 

The COLLAB dataset is a set of networks that have researchers as nodes and the edges connect those who have collaborated with each other. SD-ZFS FB was able to show a slight advantage over the greedy heuristic with a mean difference of 0.20 and a mean deviation of 0.45\%. It also showed consistent performance and rarely underperformed the greedy heuristic.

\begin{table}[h!]
\centering
\caption{SD-ZFS FB results on COLLAB dataset}
\begin{tabular}{lc}
\hline
\textbf{Outcome} & \textbf{Percentage} \\
\hline
Win     & 22.4\% \\
Win/Tie & 86.3\%  \\
Loss    & 13.7\%  \\
\hline
\end{tabular}

\label{tab:collab}
\end{table}

The REDDIT-BINARY dataset represents users as nodes and edges as comments as replies to other users. We tested SD-ZFS FB and observed slightly worse performance than greedy. This dataset was the only dataset where none of the models were able to show an improvement over the greedy heuristic. SD-ZFS BA had the best performance with a mean difference of -0.67 and a mean deviation of -0.29\%. Figure 9 shows the approximation ratio relative to the greedy solution. As the size of the graph increases, the model's performance shows improvement in solution strength relative to the greedy solution.  While the framework shows a negative deviation and difference, it still showed it was resilient by matching the greedy performance on greater than 60\% of the samples.Table 6 shows the win/loss rate.

\begin{figure}
    \caption{SD-ZFS BA Approximation to Greedy Heuristic on REDDIT dataset}
    \centering
    \includegraphics[width=1\linewidth]{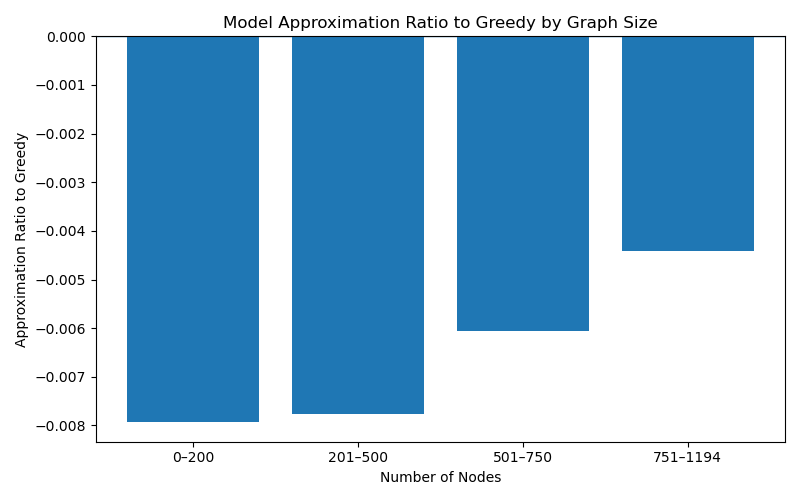}
    \label{fig:ba_approximation}
\end{figure}

\begin{table}[h!]
\caption{SD-ZFS BA results on Reddit dataset}
\centering
\begin{tabular}{lc}
\hline
\textbf{Outcome} & \textbf{Percentage} \\
\hline
Win     & 1.1\% \\
Win/Tie & 62.4\%  \\
Loss    & 37.6\%  \\
\hline
\end{tabular}
\end{table}

The IMDB-BINARY dataset represents actors and actresses as nodes and an edge is formed when two actors have acted in the same movie. All three models showed very little deviation and were either slightly better or slightly worse by a fraction of a percent. We investigated further and noted that both our model, and the greedy algorithm, closely approximated the optimal solution for this dataset. We utilized the Wavefront Algorithm to calculate the optimal solution size for all networks with less than or equal to 36 nodes(94.8\% of the networks within this dataset). SD-ZFS ER found the optimal solution with 100\% success rate and the greedy algorithm found the optimal solution with a 99.7\% success rate. On the full IMDB dataset(including all networks with greater than 36 nodes), the model showed a negligible deviation from the greedy solutions where it outperformed greedy on 6 samples and underperformed on 3 samples. 

In conclusion, for all three network families there was a model that exceeded the performance of the greedy heuristic with the exception of the REDDIT dataset where each model closely approximated the greedy solution.  In the following section we will examine the characteristics of each dataset to derive further insight into which network properties correlate to the success or failure of a model relative to the greedy solution. 

\subsection{Influence of Network Structure}

In Table 9, various network properties are shown that relate to the mean of each dataset utilized. With the exception of Average Core(Std.), each of the columns refer to the mean value from the dataset referenced. In the following paragraphs we will highlight some of the key network properties that are relevant to the observed results of each of the test datasets.

Random Networks will tend to have a low number of hubs and neutral assortativity. The expectation for these networks is a relative consistency in structure throughout the graph, and similar structure between graphs. The only key variable that changes is density which operates as a function of the p-value to generate the network. This is evidenced by the low degree gini value, assortativity, and degree skew within Table 9 for each of the ER datasets. SD-ZFS ER shows improvement over the greedy heuristic on each of these datasets indicating that it may be picking up on inherent topological properties of random networks to generate a strategy that takes advantage of these properties.

Scale-free networks follow a power law distribution meaning there will be a large number of low-degree nodes and then a long tail of high-degree nodes that represent hubs in the network. The power law distribution leads to a consistent structure as the networks grow. This distribution leads to the networks to tend towards disassortativity and higher degree gini as shown in Table 9. Similar to SD-ZFS ER, SD-ZFS BA outperforms greedy on the test datasets where the networks are similar in structure. This shows a further indication that the S2V-DQN framework captures the structure of the network and is able to exploit the inherent properties of scale-free networks to outperform greedy when given similar test data. 

The COLLAB dataset is a real-world dataset and shares properties with scale-free networks such as a higher degree standard deviation. The high average clustering coefficient indicates the networks may contain a significant number of hubs that would be far greater than what would be observed in a random network. Additionally, the transitivity is high which is similar to the other real-world networks of Facebook and IMDB\_BINARY. This property indicates a high number of triangles throughout the network which is commonly seen in real-world data. The favorable performance of SD-ZFS BA on this dataset noted in section 4.5 indicates that the framework has the capability of picking up on some of the structure that scale-free and real-world networks share leading to an advantageous solution policy. 

The IMDB\_BINARY is unique in the way that both the framework and the greedy heuristic were able to approximate the optimal solution with a high level of precision. A couple of the key properties that IMDB exhibits are its high density, low degree standard deviation, and extremely high clustering coefficient. This would indicate that the graph is highly connected and neighbors are often connected to each other forming cliques. As mentioned previously, complete graphs have $Z(G)= |V-1|$\cite{aim2008zfs}. When components of a network have a density that nears completeness, we can expect to see $Z(G)$ approch $|V|$.

For the networks of 36 nodes or less within this dataset(948/1000 networks), the average number of nodes is 18.09 and the average optimal solution size is 14.84. This data indicates that on average for this dataset, the optimal solution is close to the size of the network. This converges with the fact that as the density and clustering coefficient of a network rise, the minimum ZFS set gets closer to the total number of vertices(See Figures 11 and 12). 

When networks are this densely connected with a high clustering coefficient, it seems that the strategy to find the minimum ZFS is largely inconsequential. The most computationally efficient strategy in these cases could be random selection of nodes if an optimal solution isn't required. Figures 11 and 12 show how the greedy algorithm performs on 100 node networks as the density and average clustering coefficient rise. 

Lastly, the REDDIT dataset has some properties that distinguish it from the rest of the datasets. There is the highest degree skew out of all of the datasets. Further, the average degree, density, transitivity, and average core standard deviation are exceptionally low. The dataset also tends towards disassortativity. When taken together, these properties indicate many of the networks in this dataset are similar to a "hub-and-spoke" where there is one central node and many leaves. This was further confirmed when analyzed in the context of the number of leaves to number of nodes. In this dataset, the average fraction of leaves is 61\%. This data point combined with low transitivity is indicative of a large number of hub-and-spoke networks. As shown in the below proof, a hub-and-spoke graph has an optimal minimum ZFS $Z(G)= |V|-2$. The unique structure of this dataset likely contributed to the result observed in section 4.5 where the framework slightly underperformed the greedy heuristic. Similar to the results observed in the IMDB dataset, it may be beneficial for networks that closely resemble hub-and-spoke networks to be solved through random vertex selection rather than any computational algorithm or machine learning framework. 

\begin{proof}
Let $G$ be a hub-and-spoke graph with hub vertex $h$ and leaf vertices
$L = V \setminus \{h\}$. We show that the optimal minimum ZFS for a hub-and-spoke network is
$|V|-2$.\\
\textit{Case 1: The first selected vertex is the hub $h$.}\\
Suppose the first node added to the zero-forcing set is the hub $h$. Under the color-change rule(Algorithm 1), the hub can only force one of its leaves if it is the only uncolored leaf. Therefore, a new leaf must be added to the minimum zero-forcing set. By definition of a hub-and-spoke graph, the only neighbor of the leaf is the hub which is already blue. Thus, the process continues until every leaf with the exception of one is added to the zero-forcing set and we derive $Z(G) = |V|-1$.\\
\textit{Case 2: The first selected vertex is a leaf.}\\
Instead, suppose the first node added to the zero-forcing set is a leaf. This leaf will only have one neighbor, $h$, which is uncolored. The color change rule forces this node blue. The process halts unless there is only one additional uncolored leaf. The next selected node must be either the hub, $h$, or another leaf. If the hub is selected, then you have the same structure shown in case 1 where the hub and a leaf have been added to the minimum ZFS and the process continues until $Z(G) = |V|-1$. If a hub is not selected and leaves are continually added, eventually all of the leaves with the exception of one will be in the minimum zero-forcing set. Once this state is reached, either of the two uncolored leaves force the hub through the color change rule. Subsequently, the hub can force the remaining leaf blue since it will be the only uncolored node left in the graph. Therefore, the size of the minimum ZFS when iteratively selecting leaves ends up being $|V|-2$.\\
As demonstrated in cases 1 and 2, hub $h$ can be added to the minimum ZFS resulting in a solution of $|V|-1$. If only leaves are added to $Z(G)$, the solution size is $|V|-2$ proving the optimal solution size to a hub-and-spoke network is $|V|-2$. Please refer to figure 10 as an example. 
\end{proof}

\begin{figure}
    \caption{Case 2 - Three leaves force the remaining nodes}
    \centering
    \includegraphics[width=0.75\linewidth]{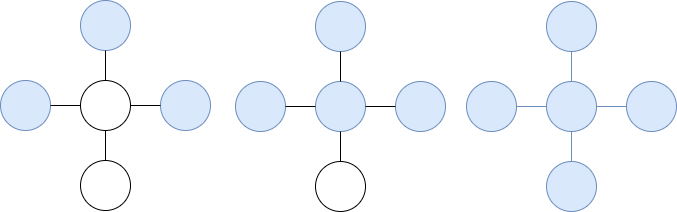}
    
    \label{fig:case2}
\end{figure}

\begin{figure}
\caption{Size of Greedy Solution vs. Network Density}
    \centering
    \includegraphics[width=1\linewidth]{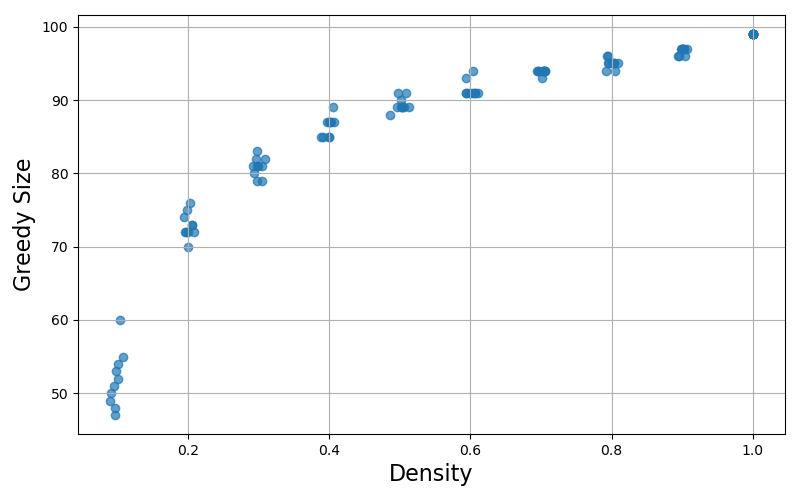}
    \label{fig:size-density}
\end{figure}

\begin{figure}
\caption{Size of Greedy Solution vs. Avg. Clustering Coefficient}
    \centering
    \includegraphics[width=1\linewidth]{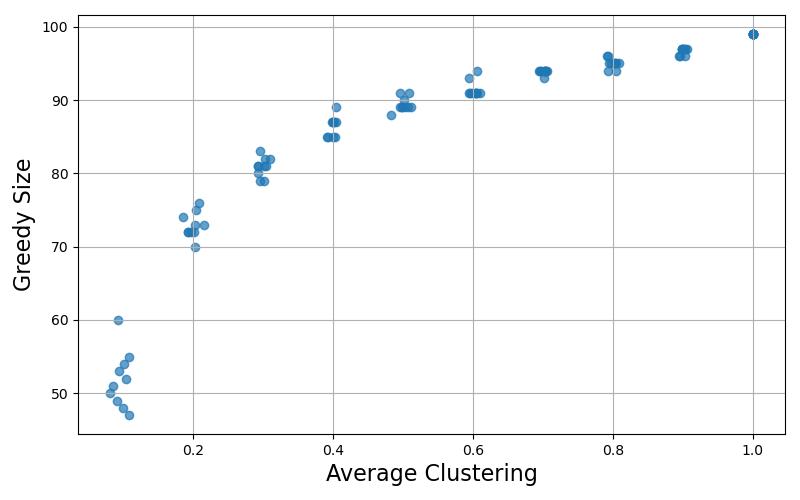}
    
    \label{fig:size-clustering}
\end{figure}

To summarize the above results, the framework outperformed the greedy heuristic on all datasets with the exception of Reddit which has a consistent unique structural property. Overall, the results provide a good point for further research into how the framework can be improved to better generalize and scale to a variety of network structures. The results are seen in Figures \ref{fig:size-density} through \ref{fig:size-clustering}.

Our results show that distinct network properties such as density, or hub-and-spoke properties, can render the $Z(G)$ computation strategy inconsequential. Further work remains on adapting the framework to address networks that have these specific configurations and also to account for trees/lines which are computationally easy to compute (Section 2.3).

\begin{table}[!h]
\centering
\caption{Performance on small graphs. }
\small
\setlength{\tabcolsep}{2.5pt}
\begin{tabular}{llrrrrrr}
\toprule
 &  & \multicolumn{3}{c}{Small ER} & \multicolumn{3}{c}{Small BA} \\
\cmidrule(lr){3-5} \cmidrule(lr){6-8}
Model &  
& Dev. & Diff. & Ratio 
& Dev. & Diff. & Ratio \\
\midrule
SD-ZFS ER &  & 11.45\% & 4.47 & 1.05 & -1.09 & -0.043 & 1.026 \\
SD-ZFS BA &  & 10.39\% & 4.06 & 1.06 & 1.017 & 0.145 & 1.021 \\
SD-ZFS FB &  & -7.85\% & -3.13 & 1.09 & -1.756 & -0.121 & 1.034 \\
\bottomrule
\end{tabular}
\label{tab:small-graphs}
\end{table}

\begin{table*}[h!]
\caption{Performance on large and real-world graphs. }
\centering
\small
\setlength{\tabcolsep}{3pt}
\begin{tabular*}{\textwidth}{@{\extracolsep{\fill}}llrrrrrrrrrr@{}}
\toprule
 &  & \multicolumn{2}{c}{Large ER} 
 & \multicolumn{2}{c}{Large BA} 
 & \multicolumn{2}{c}{COLLAB} 
 & \multicolumn{2}{c}{IMDB} 
 & \multicolumn{2}{c}{REDDIT} \\
\cmidrule(lr){3-4} \cmidrule(lr){5-6} \cmidrule(lr){7-8} \cmidrule(lr){9-10} \cmidrule(lr){11-12}
Model &    
& Dev. & Diff. 
& Dev. & Diff. 
& Dev. & Diff. 
& Dev. & Diff. 
& Dev. & Diff. \\
\midrule
SD-ZFS ER & 
& 0.71\% & 1.37 
& -0.64\% & -2.77 
& -0.30\% & -0.13 
& 0.00\% & 0.00 
& -0.95\% & -1.56 \\

SD-ZFS BA & 
& 1.59\% & 6.23 
& 1.18\% & 1.24 
& -0.02\% & -0.07 
& 0.00\% & 0.00 
& -0.29\% & -0.67 \\

SD-ZFS FB & 
& -3.06\% & -11.5 
& -2.13\% & -6.41 
& 0.45\% & 0.201 
& 0.00\% & 0.00 
& -0.80\% & -1.21 \\
\bottomrule
\end{tabular*}
\label{tab:large-graphs}
\end{table*}
\begin{table*}[t]
\caption{Structural properties of synthetic and real-world graph datasets.}
\centering
\Small
\setlength{\tabcolsep}{3pt}
\begin{tabular*}{\textwidth}{@{\extracolsep{\fill}}lrrrrrrrrrrrr@{}}
\toprule
Dataset 
& Degree Gini 
& Degree Skew 
& Degree Std 
& Avg Degree 
& Density $p$ 
& Max Core 
& Avg Core 
& Avg Core (Std) 
& Assort. 
& Avg Clust. 
& Trans. 
& Degree Bin. Wass. \\
\midrule
Large\_ER      & 0.14 & 0.24  & 4.24  & 19.64 & 0.03 & 13.82 & 13.61 & 6.48  &  0.00 & 0.03 & 0.03 & 0.22 \\
Large\_BA      & 0.30 & 2.81  & 19.13 & 29.47 & 0.10 & 15.74 & 15.69 & 8.02  & -0.06 & 0.17 & 0.16 & 9.75 \\
COLLAB         & 0.22 & 0.79  & 11.78 & 37.37 & 0.51 & 40.53 & 34.34 & 43.81 & -0.03 & 0.89 & 0.77 & 6.87 \\
IMDB\_BINARY   & 0.15 & 2.16  & 2.78  & 8.89  & 0.52 & 9.16  & 8.18  & 5.15  & -0.14 & 0.95 & 0.77 & 1.41 \\
Facebook       & 0.20 & 0.06  & 6.28  & 18.01 & 0.59 & 14.34 & 12.95 & 6.10  & -0.23 & 0.80 & 0.72 & 3.20 \\
REDDIT         & 0.47 & 12.94 & 7.37  & 2.35  & 0.02 & 2.30  & 1.30  & 0.17  & -0.36 & 0.05 & 0.01 & 1.38 \\
Scale\_BA      & 0.34 & 3.89  & 6.89  & 7.77  & 0.05 & 4.00  & 3.99  & 0.03  & -0.11 & 0.12 & 0.09 & 2.46 \\
Scale\_ER      & 0.08 & 0.11  & 7.08  & 75.72 & 0.20 & 62.32 & 62.18 & 48.89 & -0.02 & 0.20 & 0.20 & 0.58 \\
Small\_ER      & 0.14 & 0.19  & 3.10  & 12.67 & 0.20 & 8.90  & 8.69  & 1.49  & -0.04 & 0.20 & 0.20 & 0.44 \\
Small\_BA      & 0.34 & 2.19  & 2.07  & 2.78  & 0.14 & 1.51  & 1.50  & 0.50  & -0.37 & 0.16 & 0.09 & 1.14 \\
\bottomrule
\end{tabular*}
\label{tab:graph-properties}
\end{table*}

\subsection{Performance Comparison to GCN}
\begin{table}[h!]
\centering
\caption{Performance comparison on Small ER and IMDB Binary datasets. Dev. denotes mean deviation and Diff. denotes mean difference.}
\scriptsize
\setlength{\tabcolsep}{3pt}
\begin{tabular}{lrrrr}
\toprule
 & \multicolumn{2}{c}{Small ER} & \multicolumn{2}{c}{IMDB Binary} \\
\cmidrule(lr){2-3} \cmidrule(lr){4-5}
Framework 
& Dev. & Diff. 
& Dev. & Diff. \\
\midrule
SD-ZFS ER & 11.45\% & 4.47 & 0.00\% & 0.00 \\
GCN     & 5.36\%  & 2.06 & 0.00\% & 0.00 \\
\bottomrule
\end{tabular}
\label{tab:small_er_imdb}
\end{table}
In this section we compare our approach to the GCN approach taken by \citet{ahmad2024gml}. As mentioned in section 2.4, their research utilizes a combination of random selection and a GCN to derive a minimum zero-forcing set from a network. We followed their approach and trained their GCN framework with data from the ER\_Small dataset and utilize a maximum degree of 32 degrees. We utilize the model to compare the accuracy between the GCN and the S2V-DQN on two sets of training data: ER\_Small and IMDB\_BINARY. 

SD-ZFS ER is trained on randomized Erdős–Rényi networks which are similar to the data in Small\_ER. We utilize this model to compare performance between the two architectures which have been trained on similar data. As shown in Table 10,  SD-ZFS ER showed an increase in solution quality over GCN for the Small\_ER dataset. Both frameworks closely approximate greedy and optimal on the IMDB dataset which can be explained by the network properties discussed in the prior section. 

\section{Conclusion}
In conclusion, we have proposed an adapted framework of S2V-DQN to the ZFS problem(SD-ZFS) where we take a machine-learning approach to finding the minimum zero-forcing set. The framework showed success across various datasets. We demonstrated the capability of the framework to generalize and scale to unseen data that differed in structure. We have also analyzed various network properties that influence the difficulty of finding the minimum $Z(G)$. In the future we aim to adjust the framework to better account for specific network topological properties that can increase or decrease difficulty such as the presence of a complete component or hub-and-spoke, and the presence of trees or lines.

\section*{Reproducibility}

Data and code produced in this study are available upon request.

\bibliographystyle{ACM-Reference-Format}
\bibliography{references}

\end{document}